\documentclass[letterpaper]{article} 
\usepackage{aaai2026}  
\usepackage{times}  
\usepackage{helvet}  
\usepackage{courier}  
\usepackage[hyphens]{url}  
\usepackage{graphicx} 
\usepackage{natbib}  
\usepackage{caption} 
\usepackage{algorithm}
\usepackage{algorithmic}

\usepackage{amssymb}
\usepackage{booktabs}
\usepackage{multirow}
\usepackage{xcolor}

\usepackage[utf8]{inputenc}
\usepackage[most]{tcolorbox}
\tcbuselibrary{breakable}

\usepackage[table]{xcolor}
\definecolor{mygray}{gray}{.9}

\newcolumntype{C}[1]{>{\centering\arraybackslash}p{#1}}

\usepackage{listings}
\definecolor{bgcolor}{RGB}{250,250,250}
\definecolor{kw1}{RGB}{0, 0, 180}         
\definecolor{kw2}{RGB}{86,156,214}        
\definecolor{kw3}{RGB}{184,82,162}        
\definecolor{strcolor}{RGB}{206,145,120}  
\definecolor{comcolor}{RGB}{106,153,85}   
\definecolor{selfcolor}{RGB}{128,64,0}    
\definecolor{blackborder}{RGB}{0,0,0}
\definecolor{bordergray}{RGB}{80,80,80}     
\definecolor{numgray}{RGB}{110,110,110}     

\lstdefinelanguage{MyPython}{
  language=Python,
  morekeywords={[1]class,def,return},
  morekeywords={[2]import,from,as},
  morekeywords={[3]if,else,elif,for,while,in,not,and,or},
  morekeywords={[4]self,True,False,None},
  sensitive=true
}

\lstdefinestyle{vscodeLight}{
  backgroundcolor=\color{bgcolor},
  basicstyle=\ttfamily\small,
  keywordstyle={[1]\color{kw1}\bfseries},
  keywordstyle={[2]\color{kw2}},
  keywordstyle={[3]\color{kw3}},
  keywordstyle={[4]\color{selfcolor}\bfseries},
  commentstyle=\color{comcolor}\itshape,
  stringstyle=\color{strcolor},
  numberstyle=\tiny\color{numgray},
  numbers=left,
  numbersep=9pt,
  frame=single,                
  rulecolor=\color{bordergray},     
  framesep=5pt,               
  framerule=1.2pt,             
  tabsize=4,
  breaklines=true,
  showstringspaces=false,
  captionpos=b,
  columns=flexible
}

\usepackage{newfloat}
\usepackage{listings}
\DeclareCaptionStyle{ruled}{labelfont=normalfont,labelsep=colon,strut=off} 
\floatstyle{ruled}
\newfloat{listing}{tb}{lst}{}
\floatname{listing}{Listing}
\usepackage{amsmath}
\usepackage{bm}
\usepackage{amssymb}
\usepackage{booktabs}
\usepackage{multirow}
\usepackage{xcolor}
\usepackage{enumitem}
\usepackage{bibentry}
\usepackage[hang,flushmargin]{footmisc}

\title{\textbf{A\textsuperscript{2}Flow}: Automating Agentic Workflow Generation via Self-Adaptive\\Abstraction Operators}
\author{
    Mingming Zhao\textsuperscript{\rm 1},
    Xiaokang Wei\textsuperscript{\rm 1,\rm 2}\thanks{Corresponding author.},
    Yuanqi Shao\textsuperscript{\rm 1,\rm 3},\\
    Kaiwen Zhou\textsuperscript{\rm 1,\rm 4},
    Lin Yang\textsuperscript{\rm 1},
    Siwei Rao\textsuperscript{\rm 5},
    Junhui Zhan\textsuperscript{\rm 5},
    Zhitang Chen\textsuperscript{\rm 1},
}
\affiliations{
    \textsuperscript{\rm 1}	Huawei Noah’s Ark Lab, \\
    \textsuperscript{\rm 2}The Hong Kong Polytechnic University, \\
    \textsuperscript{\rm 3}State Key Laboratory of Multimodal Artificial Intelligence Systems, CASIA,\\
    \textsuperscript{\rm 4} The Chinese University of Hong Kong, 
    \textsuperscript{\rm 5} Huawei Technologies Co., Ltd \\
    zhaomingming9@huawei.com\\ 
}

\usepackage{bibentry}

\begin{document}

\maketitle

\begin{abstract}
Large language models (LLMs) have shown strong potential in automating the design of agentic workflows. However, existing methods still rely heavily on manually predefined operators, limiting generalization and scalability. To address this issue, we propose \textbf{A\textsuperscript{2}Flow}, a fully automated framework for agentic workflow generation based on \textit{self-adaptive abstraction operators}. \textbf{A\textsuperscript{2}Flow} employs a three-stage operator extraction process: 1) Case-based Initial Operator Generation: leveraging expert demonstrations and LLM reasoning to generate case-specific operators; 2) Operator Clustering and Preliminary Abstraction: grouping similar operators across tasks to form preliminary abstractions; and 3) Deep Extraction for Abstract Execution Operators: applying long chain-of-thought prompting and multi-path reasoning to derive compact and generalizable execution operators. These operators serve as reusable building blocks for workflow construction without manual predefinition. Furthermore, we enhance node-level workflow search with an \textit{operator memory mechanism}, which retains historical outputs to enrich context and improve decision-making. Experiments on general and embodied benchmarks show that \textbf{A\textsuperscript{2}Flow} achieves a 2.4\% and 19.3\% average performance improvement and reduces resource usage by 37\% over state-of-the-art baselines.
\end{abstract}

\begin{links}
    \link{Code}{https://github.com/pandawei-ele/A2FLOW}
\end{links}

\section{Introduction}

Large Language Models (LLMs) have demonstrated remarkable capabilities across diverse domains, including code generation, data analysis, decision-making, question answering, autonomous driving, and even embodied-ai reasoning~\cite{zhu2024autotqa,zhuge2024gptswarm,liu2024survey,xie2024haichart,zhong2024achieving,cao2025pgpo,madaan2023self}. The emergence of LLM agents—autonomous systems that leverage LLMs to perceive, reason, and act further extends their potential across complex tasks and environments. However, these agents often rely on manually crafted agentic workflows, which are structured sequences of LLM invocations guided by human-designed instructions~\cite{brown2020language}. 


\begin{figure}[ht]
  \centering
  \includegraphics[width=1\linewidth]{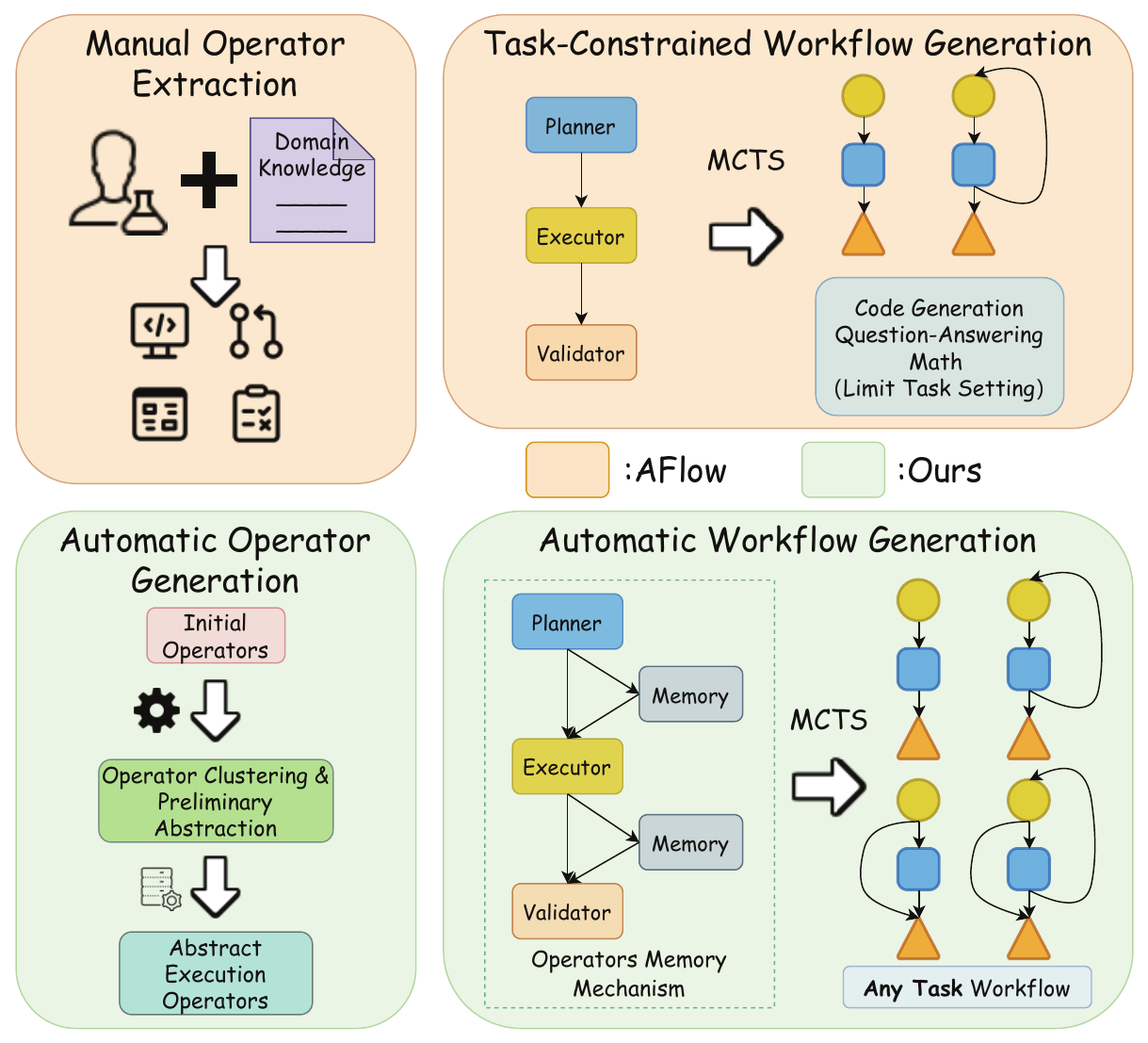}
  \caption{Compared with existing methods, A\textsuperscript{2}Flow improves performance by: (1) Self-Adaptive Abstraction Operators module that optimizes the search process and improves efficiency, and (2) Operators Memory Mechanism that leverages enables more context-aware execution and improving workflow search performance.}
  \label{fig:teaser}
\end{figure}
The process of designing and refining such workflows, requiring substantial human effort, constrains the scalability of LLMs, their adaptability to novel domains, and their generalization ability across diverse tasks~\cite{tang2023verifai,zhang2410aflow, xietravelplanner,wang2024tdag}.

Recent research seeks to automate the generation and optimization of effective agentic workflows. DSPy~\cite{khattab2024dspy} introduces a structured framework that formalizes language model workflows as learnable and optimizable text transformation graphs, enabling automated construction and tuning of LLM pipelines without relying on manually crafted prompt templates.
ADAS~\cite{hu2024automated} designs agentic systems through code-based workflows, but its linear heuristic search mechanism lacks efficiency, making it difficult to produce optimal workflows within a restricted number of iterations.

AFLOW~\cite{zhang2410aflow} introduces an automated workflow generation framework that formulates workflow optimization as a code-based search problem and efficiently explores he vast search space using a variant of MCTS. It introduces Operators: reusable bundles that wrap common agent actions (e.g., Ensemble, Review, Revise) by packaging nodes and edges into a simple interface. Given a predefined set of Operators and the edge space, AFLOW can build and refine workflows more efficiently. 

However, AFLOW still depends on manually crafted, task‑specific operators, which limit the automation of workflow construction and hinder generalization to open-world or embodied tasks, and cannot guarantee optimality within the search space, as illustrated at the top of Figure
DebFlow~\cite{su2025debflow} leverages a debate mechanism and integrates reflection to optimize workflows and reduce resource usage; however, it still relies on the same predefined operators as AFLOW, making it prone to redundant operators and unnecessary overhead.
MermaidFlow~\cite{zheng2025mermaidflow} proposes domain-aware evolutionary operators to preserve semantic correctness while promoting structural diversity. Nevertheless, due to its predefined operator initialization design, it still struggles to generalize to certain open-world tasks, such as Embodied Scenarios and Everyday Scenarios. Despite these advancements, full automation has not been achieved.

To address the limitations of existing agentic workflows and realize fully automated workflow generation, we introduce \textbf{A\textsuperscript{2}Flow}, an agentic workflow generation framework equipped with \textit{Self-Adaptive Abstraction Operators} to optimize the search process and improve efficiency. 
By removing the dependency on manually predefined, task-specific operators, A\textsuperscript{2}Flow reduces reliance on human expertise in workflow design and mitigates search inefficiencies caused by manually introduced biases, thereby enhancing the scalability, adaptability, and generalization ability of LLMs across diverse and complex tasks.
Additionally, we integrate an \textit{Operators Memory Mechanism} that reuses agents' thought processes, reducing interaction steps during inference and improving generation efficiency. Specifically, A\textsuperscript{2}Flow is designed to autonomously derive optimal abstract operators directly from raw expert demonstrations, eliminating manual predefinition through a novel pipeline (shown at the bottom of Figure~\ref{fig:teaser}). This process consists of three key stages:
1) \textit{Initial Operator Generation}: Using expert data and the LLM, prompts generate case-aware operators for each task type; 2) \textit{Operator Clustering and Abstraction}: Operators with similar functions are clustered across cases to form preliminary abstract operators; 3) \textit{Operator Refinement and Selection}: Preliminary operators undergo deep reasoning via Long Chain-of-Thought (COT)~\cite{wei2022chain} and feasibility checks, followed by a self-consistency-based multi-path search to select final task-aware abstract operators. Besides, when deploying these adaptive operators for workflow search, the \textit{Operator Memory Mechanism} enhances information sharing and strengthens task comprehension capabilities.

The key contributions of this work are as follows:
\begin{itemize}
\item We propose \textbf{A\textsuperscript{2}Flow}, a highly automated framework for agentic workflow generation that extracts \textit{Self-Adaptive Abstraction Operators} to reduce computational costs and significantly improve generalization across diverse tasks.

\item We introduce an \textit{Operators Memory Mechanism}, which augments each operator with access to accumulated historical outputs, enabling more context-aware execution and improving workflow search performance.

\item Extensive experiments on eight benchmark datasets across five distinct domains, including code generation, mathematical reasoning, reading comprehension, games, and embodied tasks, demonstrate that A\textsuperscript{2}Flow consistently outperforms state-of-the-art methods and exhibits strong generalization capabilities.
\end{itemize}

\section{Related Work}
\noindent \textbf{Agentic Workflow } 
Agentic workflows and autonomous agents represent two distinct approaches to utilizing LLMs. Autonomous agents focus on dynamic decision-making based on environmental feedback~\cite{wang2023voyager, zhuge2023mindstorms, zhang2024mobileexperts,kim2024husky,song2023llm,zhou2024language}, while agentic workflows rely on predefined multi-step procedures that are typically constructed using human expertise and iterative adjustments~\cite{wang2023unleashing, yue2024mmmu,wang2024mmlupro,zheng2023judging,xin2024promptagent}. Compared to agents that demand carefully designed action spaces and control strategies, agentic workflows are often easier to implement and more adaptable in practice. Existing workflows can be classified into general-purpose and domain-specific categories. General-purpose workflows aim to handle a broad range of tasks using simple, reusable structures~\cite{wei2022chain, wang2023unleashing, madaan2023self,liuagentbench,zhang2024agentprune}, whereas domain-specific workflows are tailored for specialized applications such as code generation~\cite{hu2024automated,ridnik2024code}, data analysis~\cite{xie2024haichart}, mathematical computation~\cite{zhong2024achieving}, and complex question answering~\cite{zhou2024symbolic}. Despite their usefulness, manually defined workflows face two main challenges: general workflows often struggle with complex task reasoning, and domain-specific workflows lack flexibility beyond their intended scope~\cite{hu2024agentgen,yao2022webshop}.

\noindent \textbf{Automated Agentic Optimization } Recent work to automate agentic workflow design fall into three categories: prompt optimization, hyperparameter tuning, and full workflow optimization~\cite{qiao2024autoact,huang2023agentcoder,zhong2024achieving,yang2024buffer,dong2024agent}. Prompt optimization~\cite{fernando2023promptbreeder,yu2023thought,sumerscognitive} improves prompts within fixed workflows using LLMs, but often lacks generalization and requires manual effort. Hyperparameter tuning~\cite{saad2024archon,liu2023dynamic,zhou2023llm} adjusts preset parameters but has a limited scope. In contrast, automated workflow optimization aims to optimize the entire workflow structure and offers greater potential for fully automatic and generalizable workflow generation.
GPTSwarm~\cite{zhuge2024gptswarm} adopts graph structures combined with reinforcement learning, but its design struggles to represent workflows involving conditional states, limiting its expressiveness.
ADAS~\cite{hu2024automated} represents workflows as code and stores past workflows in a linear list, which aligns with our goals. However, its simple experience representation limits the search efficiency, making it hard to find effective workflows. AFlow~\cite{zhang2410aflow} also uses code to represent workflows, but goes further by providing a more fundamental structure called named node. AFlow~\cite{zhang2410aflow} searches workflows with MCTS over code-represented nodes and edges, and it relies on predefined operators that bundle fixed node combinations. These operators are hand-crafted per task from experience, which lowers automation and requires redesign for each new setting. As a result, AFlow transfers poorly to unseen or open-world tasks (e.g., embodied control, WebShop), and there is no clear evidence that its operators are close to optimal. DebFlow~\cite{su2025debflow} employs a debate mechanism and integrates reflection to optimize workflows and reduce resource usage. However, it inherits AFlow's~\cite{zhang2410aflow} predefined operators, leaving it susceptible to redundant operators and unnecessary overhead.
MermaidFlow~\cite{zheng2025mermaidflow} still struggles to generalize to certain open-world tasks, such as Embodied Scenarios and Everyday Scenarios.
\textbf{A\textsuperscript{2}Flow} addresses this gap by adaptively extracting abstract operators from expert data without manual design or prior knowledge, enabling automatic discovery of effective operator sets that improve workflow automation and generalization.

\section{Method}

\begin{figure*}[tb]
  \centering
    \includegraphics[width=1\linewidth]{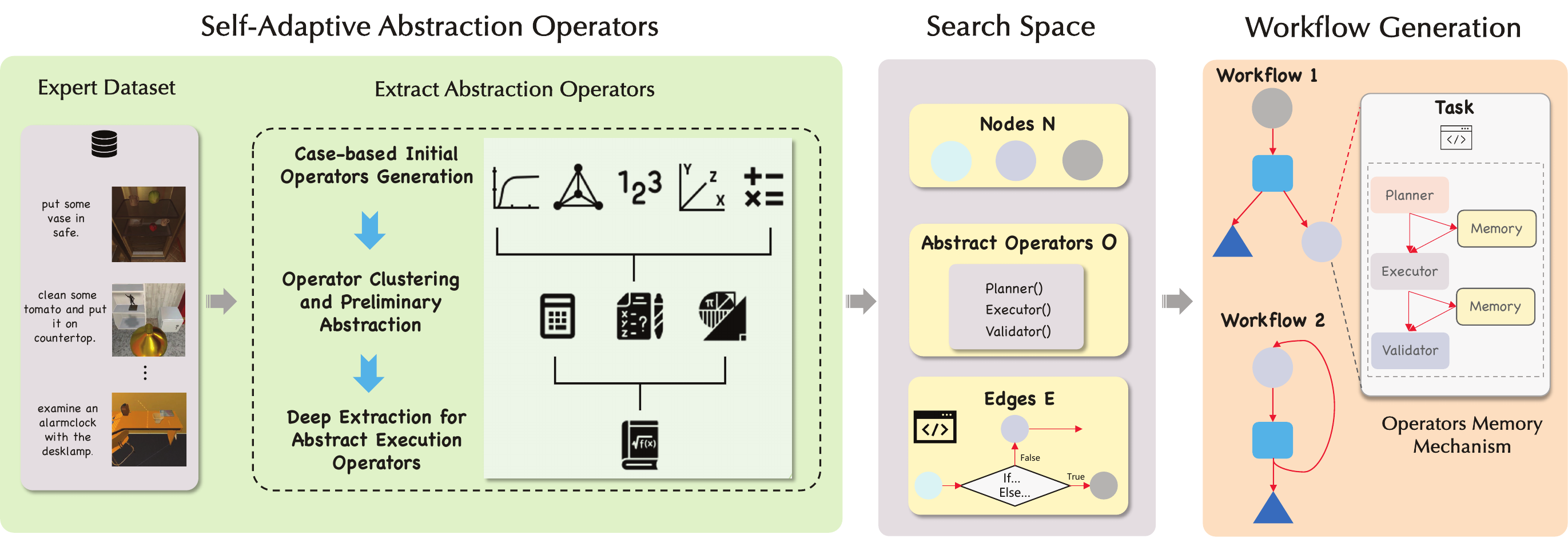}
      \caption{Overview of our framework. Left: Our method synthesizes abstract execution operators through a three-phase generation process, leveraging expert data for iterative refinement. Middle: setting a search space composed of nodes, a abstract operator set, and a code representing edge. Right: Illustration of the MCTS-based evolutionary workflow process. Through
      integrating an Operators Memory Mechanism, A\textsuperscript{2}Flow enhance the  workflow search capability at each node.}
  \label{fig:overview}
\end{figure*}

\subsection{Preliminary}

Recently, AFLOW~\cite{zhang2410aflow} exhibits strong performance and efficiency in automated agentic optimization, which leverages Large Language Models (LLMs) as optimizers within a variant of Monte Carlo Tree Search (MCTS) to search for optimal workflows. To efficiently explore this vast search space, AFLOW~\cite{zhang2410aflow} introduces Operators, which encapsulate common agentic operations (e.g., Ensemble, Review, Revise) by unifying N and E into a single interface, enabling AFlow to perform faster search and streamlined workflow generation.
Specifically, AFlow defines an \textit{agentic workflow} $\mathcal{W}$ as a series of LLM-invoking nodes connected by edges to define the execution orders, denoted as $\mathcal{N} = \{N_{1}, N_{2}, \ldots, N_{i}, \ldots\}$. Each node $N_{i}$ represents a specific operation performed by an LLM and is characterized by the following parameters.

A workflow node $N_{i}$ is defined by a language model $M$, an input prompt $P$, a temperature parameter $\tau$ controlling output randomness, and an output format $F$ (e.g., XML, JSON, Markdown, raw).

\textit{Edge} $E$ represents abstract structures defining node relationships, governing the sequence of execution. 

Given a task \(T\) and an evaluation function \(G\), workflow optimization aims to find a workflow \(\mathcal{W}\) that maximizes \(G(\mathcal{W}, T)\). AFLOW~\cite{zhang2410aflow} formulates this as a search problem where an algorithm \(A\) explores the search space \(\mathcal{S}\), which includes all possible configurations of node parameters and edge structures. To make this exploration efficient, AFLOW~\cite{zhang2410aflow} introduces \textit{Operators} that encapsulate common agentic operations (e.g., Ensemble, Review, Revise) by unifying nodes and edges into concise interfaces, enabling AFLOW~\cite{zhang2410aflow} to achieve faster search and streamlined workflow generation.

\begin{equation}
    \mathcal{S}_{\text{AFLOW}} = 
    \left\{ (P_1, \ldots, P_n, E, O_1, \ldots, O_n) \right\},
\end{equation}
\begin{equation}
    W^{*} = \text{AFLOW}(\mathcal{S}_{\text{AFLOW}}, G, T),
\end{equation}

\noindent where $\mathcal{P}$, $\mathcal{E}$, $\mathcal{O}$ representing the sets of possible prompts, output formats, and edge configurations, respectively, and $P_i$ $\in$ $\mathcal{P}$,  $E$ $\in$ $\mathcal{E}$, $O_i$ $\in$ $\mathcal{O}$.

\subsection{Overview}
Our A\textsuperscript{2}Flow enhances the automation of generating agentic workflows by employing Large Language Models (LLMs) as optimizers within \textbf{Self-Adaptive Abstraction Operators} (Figure~\ref{fig:overview}). Before initiating workflow search, we automatically extract these operators without relying on predefined human expertise. The extraction consists of three stages: (1) \textit{Case-based Initial Operator Generation}, where prompts use expert training data and LLM reasoning to generate case-aware operators for each task type; (2) \textit{Operator Clustering and Abstraction}, where operators with similar functions across cases are grouped into preliminary abstract operators; and (3) \textit{Deep Extraction for Abstract Execution Operators}, where Long CoT~\cite{wei2022chain} and self-consistency enable deep reasoning and multi-path search to refine task-aware abstract operators. To support the workflow search process, we also propose an \textit{Operator Memory Mechanism} (Sec.~\ref{sec:Memory}) that promotes information sharing across operators and improves their task comprehension.

\subsection{Self-Adaptive Abstraction Operators}
\label{sec:Operators}
We observe that recent studies rely heavily on manually predefined operators, which limits workflow automation and hinders fast generalization to open-world tasks. To address this, we propose an automatic generation approach for code-based and self-adaptive abstract operators, which directly derives highly integrated operators from raw expert data, enhancing the ability to rapidly and automatically generate workflows for diverse tasks.

\noindent \textbf{Case-based Initial Operators Generation}
Following AFLOW~\cite{zhang2410aflow}, we partition the expert dataset for the specific task into a 20\% validation set and an 80\% test set, with the random seed fixed at 42. 
Using the validation set, we design initial operator-extraction prompts and leverage the reasoning capability of the LLM to perform initial case-aware operator extraction. 
Formally, the initial operator set \(O\) is defined as
\begin{equation}
\begin{split}
O^{(e)} = \{ \, & o_{i,j} = E(C_i, P_e, M) \mid \\
                & i = 1, \dots, n; j = 1, \dots, m_i \, \},
\end{split}
\end{equation}
where $n$ denotes the total number of cases, and $m_i$ is the number of operators generated for $C_i$. $E$ is the operator extraction function that, given a specific case $C_i$, the prompt $P_e$, and LLM $M$, generates a set of initial case-aware operators $O^{(e)}$.

Notably, we design prompts $P_c$ to convert each case into a code-based workflow, where each operator in $O^{(e)}$ is defined as a basic block, similar to standard program code. Each block has a single input and output, with no intermediate jumps or branching. The details of the prompts $P_c$ as shown in Figure~\ref{fig:study_case} and Appendix.

\noindent \textbf{Operator Clustering and Preliminary Abstraction} 
Since the initial operator set $O^{(e)}$ often contains operators with overlapping or similar functionalities, we further apply a functional clustering process guided by the LLM. Formally, the preliminary abstract operator set is defined as
\begin{equation}
    O^{(a)} = \mathcal{C}(O^{(e)}, P_a, M),
\end{equation}
where $P_a$ is the clustering prompt, and $\mathcal{C}$ denotes the functional clustering function that abstracts similar operators into higher-level representations. The details of the prompts $P_a$ as shown in Figure~\ref{fig:study_case} and Appendix.

\noindent \textbf{Deep Extraction for Abstract Execution Operators}
After clustering, the number of case-based operators is reduced, but they remain insufficiently abstract and general. To achieve deeper abstraction and merge the preliminary abstract operator set $O^{(a)}$ into universal execution operators (e.g., execute, verify),  we design a \textit{multi-path parallel generation} framework with Long chain-of-thought (CoT) prompting to obtain the final task-aware abstract operators $O^{(t)}$. 
By adjusting the temperature and randomness of the LLM, we generate $m$ independent reasoning paths $\{\mathcal{P}_p \}_{p=1}^{m}$, where $m=6$ in our implementation. For each path $\mathcal{P}_p$, we perform three iterative refinement steps guided by introducing CoT prompting:

\begin{enumerate}
    \item The task instructions $I$, the prompt $P_t$, and the preliminary abstract operator set $O^{(a)}$ are provided to the LLM model to generate the initial abstract operator $o_1$.
    \item The tuple $(I, P_t, o_1)$ is fed into the LLM model with CoT prompting to produce the refined operator $o_2$.
    \item The tuple $(I, P_t, o_1, o_2)$ is used with another round of CoT reasoning to obtain the final abstract operator $o_3$.
\end{enumerate}
The final task-specific operator set is obtained by aggregating the final operators from all paths:
\begin{equation}
    O^{(t)} = \mathcal{A}_t\big( \{ o_{p,3} \}_{p=1}^{6}, P_t, M \big),
\end{equation}
where \( O^{(t)} \) denotes the final set of task-aware abstract operators, 
\(\{ o_{p,3} \}_{p=1}^{6}\) represents the final candidate operators generated from each of the six reasoning paths after three-step iterative refinement, 
\(\mathcal{A}_t(\cdot)\) refers to the LLM-based aggregation function that merges and generalizes the operators across multiple paths, 
\( P_t \) is the prompt used during the aggregation stage.

\noindent \textbf{Reflection Mechanism}  
To ensure the correctness of code-based reasoning, we incorporate a reflection mechanism inspired by self-correction and error-feedback. 
At each step, the LLM is prompted to output Python code, which is then validated by a Python executor for syntax and executability. 
If the code fails the check, the executor raises an error signal that triggers the LLM to reflect and regenerate the code until a valid result is obtained. 
This iterative correction improves the reliability of the operator extraction process.

\subsection{Operators Memory Mechanism}
\label{sec:Memory}
We enhance the workflow search capability at each node by integrating an \textit{Operators Memory Mechanism} that retains and accumulates intermediate operator outputs. 
Unlike the sequential AFLOW design, where each operator \( o_k \) only depends on the immediate output of \( o_{k-1} \), 
our mechanism introduces a memory space \( \mathcal{M}_k \) to store all previous results. 
The computation of the \( k \)-th operator is thus defined as:
\begin{equation}
    o_k = f_k(\text{input}_k, P_k, \mathcal{M}_{k-1}),
\end{equation}
where \( P_k \) is the instruction prompt and \(\mathcal{M}_{k-1}\) contains all outputs up to step \( k-1 \). In each node, $f_k$ denotes the corresponding code-based operator execution function. 
After each step, the memory is updated as:
\begin{equation}
    \mathcal{M}_{k} = \mathcal{M}_{k-1} \cup \{o_k\},
\end{equation}
By incorporating both the original task context and historical information, this design empowers the operator to perform more accurately and generalize better across tasks.

\subsection{Automated Workflow Optimization}
\label{sec:workflow}
Our workflow search and optimization mechanism builds upon the AFLOW framework, 
which follows a structured cycle of \textit{initialization, selection, expansion, evaluation, and backpropagation}. 
We start from a template workflow \( W_0 \), partition the dataset into a validation set (20\%) and a test set (80\%) with a fixed random seed, 
and employ a mixed probability selection strategy to balance exploration and exploitation. 
The LLM-based optimizer expands workflows by generating or modifying nodes, 
while each candidate workflow is executed multiple times on the validation set to compute robust performance metrics, 
which are then backpropagated to update scores and historical experience.

Unlike AFLOW, which uses predefined operators, we inject our self-adaptive abstract operator set $\{O^{(t)}\}$ (Section~\ref{sec:Operators}) and Operators Memory $M$(Section~\ref{sec:Memory}) into the workflow search. 
During the expansion phase, the LLM-based optimizer directly leverages these adaptive operators, 
which are capable of dynamically generalizing and merging execution steps, as described in our three-step extraction process. 
The final optimized workflow is defined as:
\begin{equation}
    W^{*} = \mathcal{S}\big(W_0, \{ O^{(t)} \}, G, D_V, \mathcal{M} \big),
\end{equation}
where $\mathcal{S}$ denotes the AFLOW-based search and optimization procedure, 
$\{ O^{(t)} \}$ is our self-adaptive operator set, $G$ is the evaluator, and \( D_V \) is the validation dataset. 
This integration allows the search to jointly optimize workflow structures and operator representations, 
yielding more compact and task-specific solutions.

\section{Experiments}
\subsection{Experiments Setting}
\noindent \textbf{Datasets} Our experimental evaluation employed a comprehensive suite of eight publicly available benchmarks spanning five distinct domains: 1) \textbf{code generation} (HumanEval~\cite{chen2021evaluating}, MBPP~\cite{austin2021program}), 2) \textbf{math reasoning} (GSM8K~\cite{cobbe2021training}, MATH~\cite{hendrycks2021measuring}), 3) \textbf{reading comprehension} (HotpotQA~\cite{yang2018hotpotqa}, DROP~\cite{dua2019drop}), 4) \textbf{embodied task} (ALFWorld~\cite{shridhar2020alfworld}), 5) \textbf{game} (textcraft~\cite{prasad2023adapt}). For the MATH benchmark, we adhered to established protocols (AFLOW~\cite{zhang2410aflow}), utilizing identically curated problem sets derived from four prototypical problem types at complexity level 5.  Details are provided in the Appendix.

\begin{figure}[tb]
  \centering
  \includegraphics[width=0.94\linewidth]{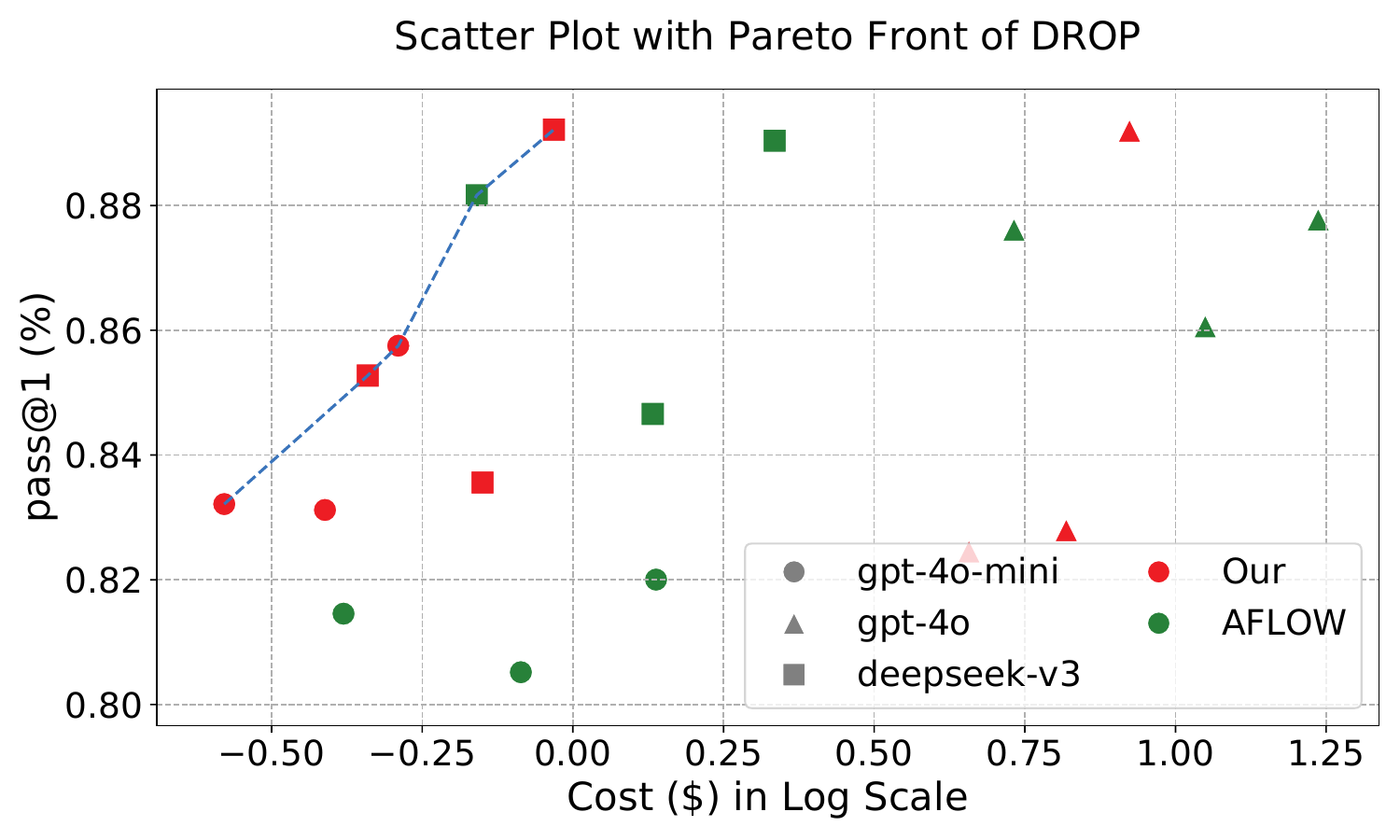}
  \caption{Total execution costs for the partitioned DROP test set are shown. A\textsuperscript{2}Flow and AFLOW-generated workflows (execution model) were evaluated using the same model. Legend colors denote the LLM executing each workflow, with exact values provided in Appendix.}
  \label{fig:cost}
\end{figure}

\noindent \textbf{Baseline} We benchmark A\textsuperscript{2}Flow against two categories of agent-based methodologies: (1) Manually engineered workflows comprising: IO(direct LLM invocation), Chain-of-Thought (CoT)~\cite{wei2022chain}, Self-Consistency (SC) with 5-answer sampling~\cite{wang2022self}, Multi-Persona debate frameworks~\cite{wang2023unleashing}, (2) Autonomous workflow optimizers:ADAS(Automated workflow synthesis via LLM agents)~\cite{hu2024automated}, AFLOW~\cite{zhang2410aflow}

\begin{table*}[!t]
  \renewcommand{\arraystretch}{1.1}
  \begin{tabular}{l c c c c c c c}
    \toprule
    \textbf{Task} & \multicolumn{2}{c}{\textbf{Reading}} &\multicolumn{2}{c}{\textbf{Coding}}&  \multicolumn{2}{c}{\textbf{Reasoning}} & \multirow{2}{*}{\textbf{Avg.}}\\
    \textbf{Method} & HotpotQA & DROP & HumanEval & MBPP & GSM8K & MATH & \\
     \midrule
      IO (GPT-4o-mini) &  68.1 & 68.3 & 87.0 & 71.8 & 92.7 & 48.6 & 72.8\\
      CoT (Wei et al., 2022) & 67.9 & 78.5 & 88.6 & 71.8 & 92.4 & 48.8 & 74.7\\
      CoT SC (5-shot) (Wang et al., 2022) & 68.9 & 78.8 & 91.6 & 73.6 & 92.7 & 50.4 & 76.0\\
      MedPrompt (Nori et al., 2023) & 68.3 & 78.0 & 91.6 & 73.6 & 90.0 & 50.0 & 75.3\\
      MultiPersona (Wang et al., 2024a) & 69.2 & 74.4 & 89.3 & 73.6 & 92.8 & 50.8 & 75.1\\
      Self Refine (Madaan et al., 2023) & 60.8 & 70.2 & 87.8 & 69.8 & 89.6 & 46.1 & 70.7\\
      \midrule
      ADAS (Hu et al., 2024) & 64.5 & 76.6 & 82.4 & 53.4 & 90.8 & 35.4 & 67.2\\
      AFLOW(Zhang et al., 2025)  & 73.5 & 80.6 & 90.9 & 83.4 & 93.5 & 56.2 & 79.6\\
      Ours  & \textbf{74.1} & \textbf{85.1} & \textbf{92.4} & \textbf{85.0} & \textbf{93.8} & \textbf{58.5} & \textbf{81.5}\\
      \bottomrule
  \end{tabular}
    \caption{We compare the performance of pre-defined operators and manually designed workflows against the fully automated framework for agentic workflow generation across reading comprehension, code generation, and mathematical reasoning scenarios. All experiments utilize the GPT-4o-mini model as executor on test set. Performance metrics are reported as the mean over three independent trials.}
  \label{tab:main_result}
\end{table*}

\begin{table}[!t]
  \centering
  \renewcommand{\arraystretch}{1.1}
  \begin{tabular}{l c c c c}
    \toprule
    \multirow{2}{*}{\textbf{Method}} & \multicolumn{2}{c}{\textbf{ALFWorld}} & \multirow{2}{*}{\textbf{TextCraft}} & \multirow{2}{*}{\textbf{Avg.}} \\
      & Seen & UnSeen &  & \\
    \midrule
      ReAct & 22.0  & 22.9 & 33.0 & 25.9 \\
      AFLOW  & 17.1 & 26.6 & 53.0 & 32.2   \\
      \midrule
      Our    & \textbf{25.0} & \textbf{31.3} & \textbf{59.0} & \textbf{38.4}  \\
      \bottomrule
  \end{tabular}
  \caption{Comparative experiments on A\textsuperscript{2}Flow vs. agentic workflow-based baselines. The best results are marked in \textbf{bold}. All experiments utilize the DeepSeek-v3 model as the executor on the test set. }
  \label{tab:emboded_result}
\end{table}

\noindent \textbf{Implementation Details} A\textsuperscript{2}Flow utilizes different models for generation, optimization. and execution. We use Claude-3.5-sonnet~\cite{sonnet2024anthropic} as the workflows' optimizer and use models: GPT-4o-mini-0718 ~\cite{gpt4omini2024openai}, GPT-4o~\cite{gpt4o2024openai} and deepseek-v3~\cite{deepseekv252024Deepseek} as executors. we use models deepseek-v3~\cite{deepseekv252024Deepseek} as abstract operations' generator.

\noindent \textbf{Evaluation Metrics} We employ distinct evaluation metrics tailored to each benchmark. ALFWorld and TextCraft are indicated via binary rewards, signifying overall success or failure. For the mathematical reasoning datasets, GSM8K and MATH $_{lv5}$, the Solve Rate serves as the primary performance measure. Code generation capability is assessed on HumanEval and MBPP using the pass@1 metric \cite{chen2021evaluating}. Performance on the question answering benchmarks, HotpotQA and DROP, is quantified using the F1 Score. To evaluate both efficiency and accuracy, token usage across datasets is tracked to estimate cost, which is then analyzed with performance metrics to construct a Pareto front illustrating the trade-offs between methods.

\begin{table}[!t]
  \centering
  \begin{tabular}{l c c}
    \toprule
     \textbf{ Operators Type} & \textbf{Score} & $\Delta$ \textbf{Score} (\%)~$\uparrow$ \\
     \midrule
      w/o Abstraction Operators & \multirow{2}{*}{56.2}   & \multirow{2}{*}{-}  \\
      \& Operators Memory       &                             & \\
      w/o Operators Memory  & 53.9 & -4.1   \\
      Our  & 58.5 & 4.1  \\
      \midrule
      w/o Initial Operators   & 49.6  & -11.7  \\
      w/o Operator Clustering  & 54.5  & -3.0 \\
      w/o Deep Extraction Operators & 51.6 & -8.2  \\
      \bottomrule
  \end{tabular}
  \caption{Ablation study w.r.t. four components: initialization operators, clustering operators, abstract execution operators, and the operator memory mechanism. }
  \label{tab:operator_memory_ablation}
\end{table}

\subsection{Experiments Results and Analysis}
\noindent \textbf{Main Result} As shown in Table~\ref{tab:main_result}, we compare A\textsuperscript{2}Flow against 8 baselines on the math reasoning, code generation and reading comprehension benchmarks. The experimental results indicate that A\textsuperscript{2}Flow achieves SOTA performance across all primary task metrics, with the sole exception of the HumanEval benchmark~\cite{chen2021evaluating}. Compared to methods that pre-defined operators in workflow, our approach outperforms them by an average margin of 2.4\%. On the DROP benchmark specifically, A\textsuperscript{2}Flow exceeds the suboptimal method AFLOW by 4.5\%. On the MATH benchmark, it surpasses the SOTA baseline AFLOW by 4.1\%. For certain benchmarks, performance is primarily constrained by the inherent limitations of abstract operator implementation. In HumanEval and MBPP benchmarks, baselines utilizing pre-defined operators (including Python interpreter invocation) establish a strong comparison point. Consequently, optimizations relying solely on abstract operators demonstrate significantly lower effectiveness. Conversely, on the GSM8K benchmark, where baseline performance is already high, abstract operator optimization yields only modest gains. We are preliminarily exploring the generalization of this approach in embodied and game tasks. Results in Table~\ref{tab:emboded_result} demonstrate a statistically significant performance improvement of 19.3\%. This enhancement primarily stems from the workflow’s adaptive suitability for target tasks, enabled by automated search algorithms that derive abstract execution operators from limited training data.

\begin{figure*}[!t]
  \centering
  \includegraphics[width=0.81\linewidth]{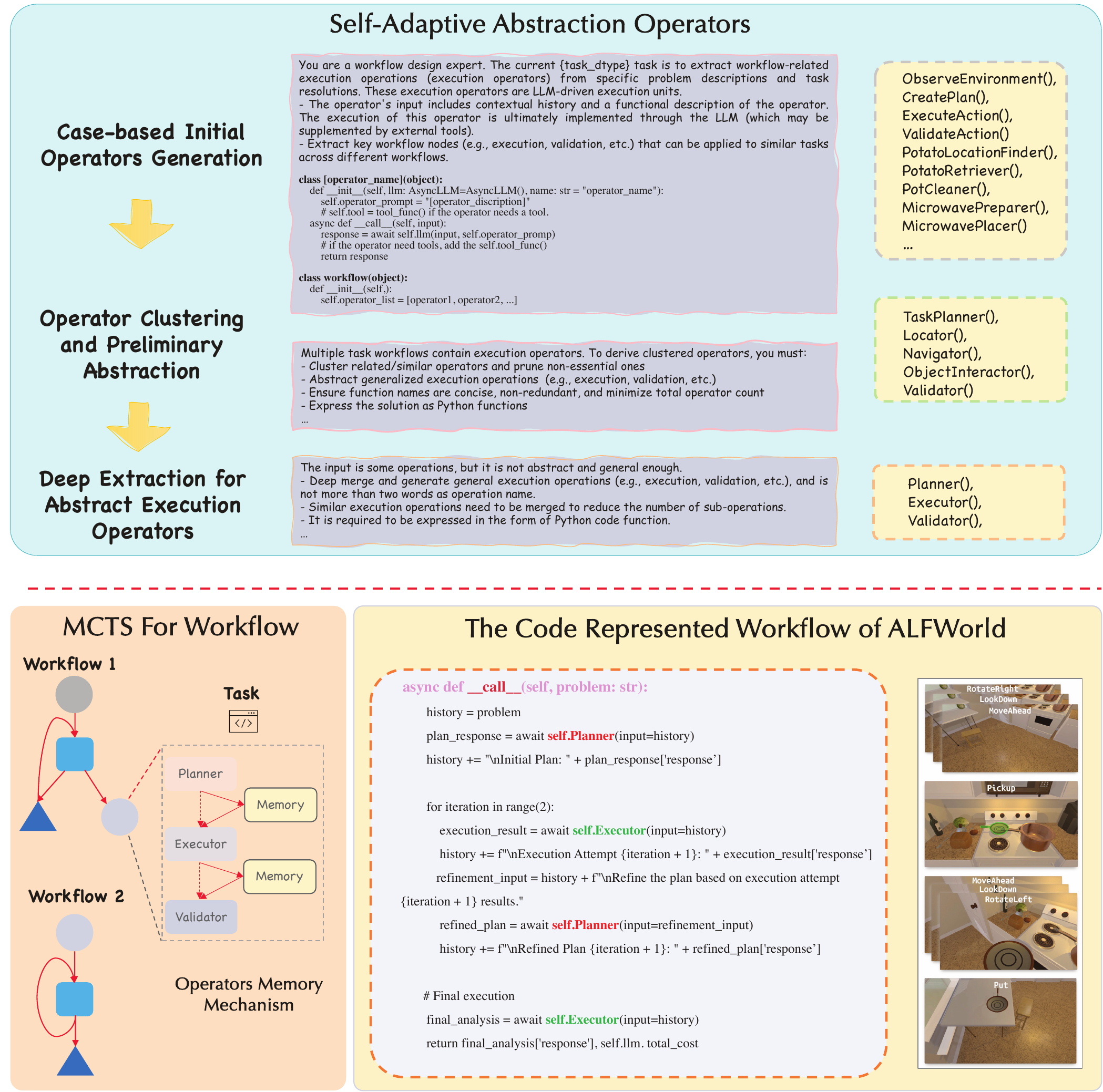}
  \caption{Self-adaptive process of operator generation and automated workflow optimization. The ALFWorld case study illustrates how A\textsuperscript{2}Flow abstracts embodied tasks into executable operators through three-phase generation, then navigates the search space via MCTS to converge on optimal structured agentic workflows.}
  \label{fig:study_case}
\end{figure*}

\noindent \textbf{Cost Analysis} 
Using GPT-4o-mini as the execution LLM, we compare the performance and cost of baseline methods with the top three workflows discovered by A\textsuperscript{2}Flow. As shown in Figure~\ref{fig:cost}, A\textsuperscript{2}Flow uncovers workflows that allow less capable models to outperform stronger ones on the cost-effectiveness Pareto front. This significantly reduces the barriers to deploying agentic workflows by automating effective workflow design and eliminating the need for manual effort. Remarkably superior cost-performance further enhances the model’s potential for widespread adoption.

\subsection{Ablation Study}
To quantify each component’s contribution, we conduct ablations on four A\textsuperscript{2}Flow variants and report both absolute \emph{Score} and \emph{$\Delta$Score} on MATH (Table~\ref{tab:operator_memory_ablation}). The full model attains 58.5\% (\,$\Delta{+}4.1$). Removing \emph{Operators Memory} reduces performance to 53.9\% (\,$\Delta{-}4.1$). Disabling the \emph{Self-Adaptive Abstraction Operators}—w/o Initial Operators, w/o Operator Clustering, and w/o Deep Extraction—yields 49.6\% (\,$\Delta{-}11.7$), 54.5\% (\,$\Delta{-}3.0$), and 51.6\% (\,$\Delta{-}8.2$), respectively. The baseline without both modules scores 56.2\%. These results show that both modules are vital, with the Self-Adaptive Abstraction Operators offering slightly larger gains.

\subsection{Case Study}
The self-adaptive process of operator generation and automated workflow optimization in A\textsuperscript{2}Flow is depicted in Figure~\ref{fig:study_case}, showing evolution from embodied task-specific samples to execution operators and workflow. In the Self-Adaptive Abstraction Operators part, for the first step, we partition the expert dataset into a 20\% validation set (33 samples). We use initial operator-extraction prompts and leverage the reasoning capability of the LLM to perform case-based initial operators generation, such as ObserveEnvironment(), CreatePlan(), etc. In the second step, we further apply a functional clustering process guided by the LLM and achieve operator clustering and preliminary abstraction. For the third step, we use a multi-path parallel generation framework with Long chain-of-thought (CoT) prompting to obtain the final alfworld embodied task abstract operators(\textit{Planner()}, \textit{Executor()}, \textit{Validator()}). In the MCTS for Workflow part, it evolves from an initial \textit{Executor()} operator to the final workflow presented. In each iteration, the optimizer dynamically integrates new operators or adjusts workflow edges to enhance performance. When a workflow contains more than two operators, node-to-node transformation messages are stored in operator memory, enabling context information sharing. Finally, we present the code for the optimal workflow discovered by A\textsuperscript{2}Flow on the ALFWorld.

\section{Conclusion}
In this work, we propose \textbf{A\textsuperscript{2}Flow}, a novel fully automated framework for agentic workflow generation via self-adaptive abstraction operators. Specifically, \textbf{A\textsuperscript{2}Flow} uses a three-step method to extract operators. It starts with case-specific operator generation, combining expert examples and LLM reasoning. Then, it performs operator clustering, grouping similar ones to form early abstractions. Finally, it applies deep reasoning with chain-of-thought prompts to refine these into compact, general-purpose execution operators. These reusable components enable workflow construction without manual definitions. A\textsuperscript{2}Flow also includes an operator memory mechanism that stores past outputs to improve node-level decision-making. Across general and embodied tasks, A\textsuperscript{2}Flow improves performance by 2.4\%, 19.3\%, and reduces resource usage by 37\% over leading baselines.

\bibliography{aaai2026}

@article{zhu2024autotqa,
  title={Autotqa: Towards autonomous tabular question answering through multi-agent large language models},
  author={Zhu, Jun-Peng and Cai, Peng and Xu, Kai and Li, Li and Sun, Yishen and Zhou, Shuai and Su, Haihuang and Tang, Liu and Liu, Qi},
  journal={Proceedings of the VLDB Endowment},
  volume={17},
  number={12},
  pages={3920--3933},
  year={2024},
  publisher={VLDB Endowment}
}

@article{zhang2410aflow,
  title={Aflow: Automating agentic workflow generation},
  author={Zhang, Jiayi and Xiang, Jinyu and Yu, Zhaoyang and Teng, Fengwei and Chen, Xionghui and Chen, Jiaqi and Zhuge, Mingchen and Cheng, Xin and Hong, Sirui and Wang, Jinlin and others},
  journal={URL https://arxiv. org/abs/2410.10762},
  year = {2024}
}

@inproceedings{khattab2024dspy,
  title={Dspy: Compiling declarative language model calls into state-of-the-art pipelines},
  author={Khattab, Omar and Singhvi, Arnav and Maheshwari, Paridhi and Zhang, Zhiyuan and Santhanam, Keshav and Haq, Saiful and Sharma, Ashutosh and Joshi, Thomas T and Moazam, Hanna and Miller, Heather and others},
  booktitle={International Conference on Learning Representations},
  year={2024}
}

@article{hu2024automated,
  title={Automated design of agentic systems},
  author={Hu, Shengran and Lu, Cong and Clune, Jeff},
  journal={arXiv preprint arXiv:2408.08435},
  year={2024}
}

@article{su2025debflow,
  title={Debflow: Automating agent creation via agent debate},
  author={Su, Jinwei and Xia, Yinghui and Shi, Ronghua and Wang, Jianhui and Huang, Jianuo and Wang, Yijin and Shi, Tianyu and Jingsong, Yang and He, Lewei},
  journal={arXiv preprint arXiv:2503.23781},
  year={2025}
}

@article{zheng2025mermaidflow,
  title={MermaidFlow: Redefining Agentic Workflow Generation via Safety-Constrained Evolutionary Programming},
  author={Zheng, Chengqi and Chen, Jianda and Lyu, Yueming and Ng, Wen Zheng Terence and Zhang, Haopeng and Ong, Yew-Soon and Tsang, Ivor and Yin, Haiyan},
  journal={arXiv preprint arXiv:2505.22967},
  year={2025}
}

@inproceedings{zhuge2024gptswarm,
  title={Gptswarm: Language agents as optimizable graphs},
  author={Zhuge, Mingchen and Wang, Wenyi and Kirsch, Louis and Faccio, Francesco and Khizbullin, Dmitrii and Schmidhuber, J{\"u}rgen},
  booktitle={International Conference on Machine Learning},
  year={2024}
}

@article{wei2022chain,
  title={Chain-of-thought prompting elicits reasoning in large language models},
  author={Wei, Jason and Wang, Xuezhi and Schuurmans, Dale and Bosma, Maarten and Xia, Fei and Chi, Ed and Le, Quoc V and Zhou, Denny and others},
  journal={Advances in Neural Information Processing Systems},
  volume={35},
  pages={24824--24837},
  year={2022}
}

@article{liu2024survey,
  title={A Survey of NL2SQL with Large Language Models: Where are we, and where are we going?},
  author={Liu, Xinyu and Shen, Shuyu and Li, Boyan and Ma, Peixian and Jiang, Runzhi and Zhang, Yuxin and Fan, Ju and Li, Guoliang and Tang, Nan and Luo, Yuyu},
  journal={arXiv preprint arXiv:2408.05109},
  year={2024}
}

@article{xie2024haichart,
  title={Haichart: Human and AI paired visualization system},
  author={Xie, Yupeng and Luo, Yuyu and Li, Guoliang and Tang, Nan},
  journal={arXiv preprint arXiv:2406.11033},
  year={2024}
}

@article{zhong2024achieving,
  title={Achieving> 97\% on gsm8k: Deeply understanding the problems makes llms better solvers for math word problems},
  author={Zhong, Qihuang and Wang, Kang and Xu, Ziyang and Liu, Juhua and Ding, Liang and Du, Bo},
  journal={arXiv preprint arXiv:2404.14963},
  year={2024}
}

@article{cao2025pgpo,
  title={PGPO: Enhancing Agent Reasoning via Pseudocode-style Planning Guided Preference Optimization},
  author={Cao, Zouying and Wang, Runze and Yang, Yifei and Ma, Xinbei and Zhu, Xiaoyong and Zheng, Bo and Zhao, Hai},
  journal={arXiv preprint arXiv:2506.01475},
  year={2025}
}

@article{tang2023verifai,
  title={VerifAI: verified generative AI},
  author={Tang, Nan and Yang, Chenyu and Fan, Ju and Cao, Lei and Luo, Yuyu and Halevy, Alon},
  journal={arXiv preprint arXiv:2307.02796},
  year={2023}
}

@article{chen2021evaluating,
  title={Evaluating large language models trained on code},
  author={Chen, Mark and Tworek, Jerry and Jun, Heewoo and Yuan, Qiming and Pinto, Henrique Ponde De Oliveira and Kaplan, Jared and Edwards, Harri and Burda, Yuri and Joseph, Nicholas and Brockman, Greg and others},
  journal={arXiv preprint arXiv:2107.03374},
  year={2021}
}

@article{austin2021program,
  title={Program synthesis with large language models},
  author={Austin, Jacob and Odena, Augustus and Nye, Maxwell and Bosma, Maarten and Michalewski, Henryk and Dohan, David and Jiang, Ellen and Cai, Carrie and Terry, Michael and Le, Quoc and others},
  journal={arXiv preprint arXiv:2108.07732},
  year={2021}
}

@article{cobbe2021training,
  title={Training verifiers to solve math word problems},
  author={Cobbe, Karl and Kosaraju, Vineet and Bavarian, Mohammad and Chen, Mark and Jun, Heewoo and Kaiser, Lukasz and Plappert, Matthias and Tworek, Jerry and Hilton, Jacob and Nakano, Reiichiro and others},
  journal={arXiv preprint arXiv:2110.14168},
  year={2021}
}

@article{yang2018hotpotqa,
  title={HotpotQA: A dataset for diverse, explainable multi-hop question answering},
  author={Yang, Zhilin and Qi, Peng and Zhang, Saizheng and Bengio, Yoshua and Cohen, William W and Salakhutdinov, Ruslan and Manning, Christopher D},
  journal={arXiv preprint arXiv:1809.09600},
  year={2018}
}

@article{dua2019drop,
  title={DROP: A reading comprehension benchmark requiring discrete reasoning over paragraphs},
  author={Dua, Dheeru and Wang, Yizhong and Dasigi, Pradeep and Stanovsky, Gabriel and Singh, Sameer and Gardner, Matt},
  journal={arXiv preprint arXiv:1903.00161},
  year={2019}
}

@article{shridhar2020alfworld,
  title={Alfworld: Aligning text and embodied environments for interactive learning},
  author={Shridhar, Mohit and Yuan, Xingdi and C{\^o}t{\'e}, Marc-Alexandre and Bisk, Yonatan and Trischler, Adam and Hausknecht, Matthew},
  journal={arXiv preprint arXiv:2010.03768},
  year={2020}
}

@article{hendrycks2021measuring,
  title={Measuring mathematical problem solving with the math dataset},
  author={Hendrycks, Dan and Burns, Collin and Kadavath, Saurav and Arora, Akul and Basart, Steven and Tang, Eric and Song, Dawn and Steinhardt, Jacob},
  journal={arXiv preprint arXiv:2103.03874},
  year={2021}
}

@article{prasad2023adapt,
  title={Adapt: As-needed decomposition and planning with language models},
  author={Prasad, Archiki and Koller, Alexander and Hartmann, Mareike and Clark, Peter and Sabharwal, Ashish and Bansal, Mohit and Khot, Tushar},
  journal={arXiv preprint arXiv:2311.05772},
  year={2023}
}

@article{wang2022self,
  title={Self-consistency improves chain of thought reasoning in language models},
  author={Wang, Xuezhi and Wei, Jason and Schuurmans, Dale and Le, Quoc and Chi, Ed and Narang, Sharan and Chowdhery, Aakanksha and Zhou, Denny},
  journal={arXiv preprint arXiv:2203.11171},
  year={2022}
}

@article{wang2023unleashing,
  title={Unleashing the emergent cognitive synergy in large language models: A task-solving agent through multi-persona self-collaboration},
  author={Wang, Zhenhailong and Mao, Shaoguang and Wu, Wenshan and Ge, Tao and Wei, Furu and Ji, Heng},
  journal={arXiv preprint arXiv:2307.05300},
  year={2023}
}

@misc{sonnet2024anthropic,
  title = {Introducing Claude 3.5 Sonnet},
  author = {{Anthropic}},
  year = {2024},
  howpublished = {\url{https://www.anthropic.com/news/claude-3-5-sonnet}},
}

@misc{gpt4omini2024openai,
  title = {{GPT-4o mini}: Advancing Cost-Efficient Intelligence},
  author = {{OpenAI}},
  year = {2024},
  howpublished = {\url{https://openai.com/index/gpt-4o-mini-advancing-cost-efficient-intelligence/}},
}

@misc{gpt4o2024openai,
  title = {Hello GPT-4o},
  author = {{OpenAI}},
  year = {2024},
  howpublished = {\url{https://openai.com/index/hello-gpt-4o/}},
}

@misc{deepseekv252024Deepseek,
  title = {{DeepSeek-V2.5}},
  author = {{Deepseek}},
  year = {2024},
  howpublished = {\url{https://huggingface.co/deepseek-ai/DeepSeek-V2.5}},
}

@article{wang2023voyager,
  title={Voyager: An open-ended embodied agent with large language models, 2023},
  author={Wang, Guanzhi and Xie, Yuqi and Jiang, Yunfan and Mandlekar, Ajay and Xiao, Chaowei and Zhu, Yuke and Fan, Linxi and Anandkumar, Anima},
  journal={URL https://arxiv. org/abs/2305.16291},
  year={2023}
}

@article{zhang2024mobileexperts,
  title={Mobileexperts: A dynamic tool-enabled agent team in mobile devices},
  author={Zhang, Jiayi and Zhao, Chuang and Zhao, Yihan and Yu, Zhaoyang and He, Ming and Fan, Jianping},
  journal={arXiv preprint arXiv:2407.03913},
  year={2024}
}

@article{zhuge2023mindstorms,
  title={Mindstorms in natural language-based societies of mind},
  author={Zhuge, Mingchen and Liu, Haozhe and Faccio, Francesco and Ashley, Dylan R and Csord{\'a}s, R{\'o}bert and Gopalakrishnan, Anand and Hamdi, Abdullah and Hammoud, Hasan Abed Al Kader and Herrmann, Vincent and Irie, Kazuki and others},
  journal={arXiv preprint arXiv:2305.17066},
  year={2023}
}

@article{madaan2023self,
  title={Self-refine: Iterative refinement with self-feedback},
  author={Madaan, Aman and Tandon, Niket and Gupta, Prakhar and Hallinan, Skyler and Gao, Luyu and Wiegreffe, Sarah and Alon, Uri and Dziri, Nouha and Prabhumoye, Shrimai and Yang, Yiming and others},
  journal={Advances in Neural Information Processing Systems},
  volume={36},
  pages={46534--46594},
  year={2023}
}

@article{zhou2024symbolic,
  title={Symbolic learning enables self-evolving agents},
  author={Zhou, Wangchunshu and Ou, Yixin and Ding, Shengwei and Li, Long and Wu, Jialong and Wang, Tiannan and Chen, Jiamin and Wang, Shuai and Xu, Xiaohua and Zhang, Ningyu and others},
  journal={arXiv preprint arXiv:2406.18532},
  year={2024}
}

@article{fernando2023promptbreeder,
  title={Promptbreeder: Self-referential self-improvement via prompt evolution},
  author={Fernando, Chrisantha and Banarse, Dylan and Michalewski, Henryk and Osindero, Simon and Rockt{\"a}schel, Tim},
  journal={arXiv preprint arXiv:2309.16797},
  year={2023}
}

@article{saad2024archon,
  title={Archon: An architecture search framework for inference-time techniques},
  author={Saad-Falcon, Jon and Lafuente, Adrian Gamarra and Natarajan, Shlok and Maru, Nahum and Todorov, Hristo and Guha, Etash and Buchanan, E Kelly and Chen, Mayee and Guha, Neel and R{\'e}, Christopher and others},
  journal={arXiv preprint arXiv:2409.15254},
  year={2024}
}

@article{brown2020language,
  title={Language models are few-shot learners},
  author={Brown, Tom B},
  journal={arXiv preprint arXiv:2005.14165},
  year={2020}
}

@inproceedings{xietravelplanner,
  title={TravelPlanner: A Benchmark for Real-World Planning with Language Agents},
  author={Xie, Jian and Zhang, Kai and Chen, Jiangjie and Zhu, Tinghui and Lou, Renze and Tian, Yuandong and Xiao, Yanghua and Su, Yu},
  year={2024},
  booktitle={Forty-first International Conference on Machine Learning}
}

@article{wang2024tdag,
  title={Tdag: A multi-agent framework based on dynamic task decomposition and agent generation},
  author={Wang, Yaoxiang and Wu, Zhiyong and Yao, Junfeng and Su, Jinsong},
  journal={arXiv preprint arXiv:2402.10178},
  year={2024}
}

@article{kim2024husky,
  title={Husky: A Unified, Open-Source Language Agent for Multi-Step Reasoning},
  author={Kim, Joongwon and Paranjape, Bhargavi and Khot, Tushar and Hajishirzi, Hannaneh},
  journal={arXiv preprint arXiv:2406.06469},
  year={2024}
}

@inproceedings{song2023llm,
  title={LLM-Planner: Few-Shot Grounded Planning for Embodied Agents with Large Language Models},
  author={Song, Chan Hee and Sadler, Brian M and Wu, Jiaman and Chao, Wei-Lun and Washington, Clayton and Su, Yu},
  booktitle={2023 IEEE/CVF International Conference on Computer Vision (ICCV)},
  pages={2986--2997},
  year={2023},
  organization={IEEE Computer Society}
}

@inproceedings{yue2024mmmu,
  title={Mmmu: A massive multi-discipline multimodal understanding and reasoning benchmark for expert agi},
  author={Yue, Xiang and Ni, Yuansheng and Zhang, Kai and Zheng, Tianyu and Liu, Ruoqi and Zhang, Ge and Stevens, Samuel and Jiang, Dongfu and Ren, Weiming and Sun, Yuxuan and others},
  booktitle={Proceedings of the IEEE/CVF Conference on Computer Vision and Pattern Recognition},
  pages={9556--9567},
  year={2024}
}

@article{wang2024mmlupro,
  title={Mmlu-pro: A more robust and challenging multi-task language understanding benchmark},
  author={Wang, Yubo and Ma, Xueguang and Zhang, Ge and Ni, Yuansheng and Chandra, Abhranil and Guo, Shiguang and Ren, Weiming and Arulraj, Aaran and He, Xuan and Jiang, Ziyan and others},
  journal={arXiv preprint arXiv:2406.01574},
  year={2024}
}

@inproceedings{liuagentbench,
  title={AgentBench: Evaluating LLMs as Agents},
  author={Liu, Xiao and Yu, Hao and Zhang, Hanchen and Xu, Yifan and Lei, Xuanyu and Lai, Hanyu and Gu, Yu and Ding, Hangliang and Men, Kaiwen and Yang, Kejuan and others},
  booktitle={International Conference on Learning Representations},
  year={2024}
}

@article{yao2022webshop,
  title={Webshop: Towards scalable real-world web interaction with grounded language agents},
  author={Yao, Shunyu and Chen, Howard and Yang, John and Narasimhan, Karthik},
  journal={Advances in Neural Information Processing Systems},
  volume={35},
  pages={20744--20757},
  year={2022}
}

@article{hu2024agentgen,
  title={AgentGen: Enhancing Planning Abilities for Large Language Model based Agent via Environment and Task Generation},
  author={Hu, Mengkang and Zhao, Pu and Xu, Can and Sun, Qingfeng and Lou, Jianguang and Lin, Qingwei and Luo, Ping and Rajmohan, Saravan and Zhang, Dongmei},
  journal={arXiv preprint arXiv:2408.00764},
  year={2024}
}

@article{ridnik2024code,
  title={Code generation with alphacodium: From prompt engineering to flow engineering},
  author={Ridnik, Tal and Kredo, Dedy and Friedman, Itamar},
  journal={arXiv preprint arXiv:2401.08500},
  year={2024}
}

@inproceedings{qiao2024autoact,
  title={Autoact: Automatic agent learning from scratch via self-planning},
  author={Qiao, Shuofei and Zhang, Ningyu and Fang, Runnan and Luo, Yujie and Zhou, Wangchunshu and Jiang, Yuchen Eleanor and Lv, Chengfei and Chen, Huajun},
  booktitle={Proceedings of the 62nd Annual Meeting of the Association for Computational Linguistics},
  year={2024}
}

@article{huang2023agentcoder,
  title={Agentcoder: Multi-agent-based code generation with iterative testing and optimisation},
  author={Huang, Dong and Bu, Qingwen and Zhang, Jie M and Luck, Michael and Cui, Heming},
  journal={arXiv preprint arXiv:2312.13010},
  year={2023}
}

@inproceedings{yu2023thought,
  title={THOUGHT PROPAGATION: AN ANALOGICAL APPROACH TO COMPLEX REASONING WITH LARGE LANGUAGE MODELS},
  author={Yu, Junchi and He, Ran and Ying, Zhitao},
  booktitle={International Conference on Learning Representations},
  year={2023}
}

@article{yang2024buffer,
  title={Buffer of Thoughts: Thought-Augmented Reasoning with Large Language Models},
  author={Yang, Ling and Yu, Zhaochen and Zhang, Tianjun and Cao, Shiyi and Xu, Minkai and Zhang, Wentao and Gonzalez, Joseph E and Cui, Bin},
  journal={arXiv preprint arXiv:2406.04271},
  year={2024}
}

@inproceedings{zhou2024language,
  title={Language Agent Tree Search Unifies Reasoning, Acting, and Planning in Language Models},
  author={Zhou, Andy and Yan, Kai and Shlapentokh-Rothman, Michal and Wang, Haohan and Wang, Yu-Xiong},
  year={2024},
  booktitle={International Conference on Machine Learning}
}

@article{zheng2023judging,
  title={Judging llm-as-a-judge with mt-bench and chatbot arena},
  author={Zheng, Lianmin and Chiang, Wei-Lin and Sheng, Ying and Zhuang, Siyuan and Wu, Zhanghao and Zhuang, Yonghao and Lin, Zi and Li, Zhuohan and Li, Dacheng and Xing, Eric and others},
  journal={Advances in Neural Information Processing Systems},
  volume={36},
  pages={46595--46623},
  year={2023}
}

@article{zhang2024agentprune,
  title={Cut the crap: An economical communication pipeline for llm-based multi-agent systems},
  author={Zhang, Guibin and Yue, Yanwei and Li, Zhixun and Yun, Sukwon and Wan, Guancheng and Wang, Kun and Cheng, Dawei and Yu, Jeffrey Xu and Chen, Tianlong},
  journal={arXiv preprint arXiv:2410.02506},
  year={2024}
}

@article{sumerscognitive,
  title={Cognitive Architectures for Language Agents},
  author={Sumers, Theodore and Yao, Shunyu and Narasimhan, Karthik and Griffiths, Thomas},
  journal={Transactions on Machine Learning Research},
  year = {2023}
}

@article{liu2023dynamic,
  title={Dynamic llm-agent network: An llm-agent collaboration framework with agent team optimization},
  author={Liu, Zijun and Zhang, Yanzhe and Li, Peng and Liu, Yang and Yang, Diyi},
  journal={arXiv preprint arXiv:2310.02170},
  year={2023}
}

@article{zhou2023llm,
  title={Llm as dba},
  author={Zhou, Xuanhe and Li, Guoliang and Liu, Zhiyuan},
  journal={arXiv preprint arXiv:2308.05481},
  year={2023}
}

@article{dong2024agent,
  title={Agentcoder: Multi-agent-based code generation with iterative testing and optimisation},
  author={Huang, Dong and Zhang, Jie M and Luck, Michael and Bu, Qingwen and Qing, Yuhao and Cui, Heming},
  journal={arXiv preprint arXiv:2312.13010},
  year={2023}
}

@inproceedings{xin2024promptagent,
  author       = {Xinyuan Wang and
                  Chenxi Li and
                  Zhen Wang and
                  Fan Bai and
                  Haotian Luo and
                  Jiayou Zhang and
                  Nebojsa Jojic and
                  Eric P. Xing and
                  Zhiting Hu},
  title        = {PromptAgent: Strategic Planning with Language Models Enables Expert-level Prompt Optimization},
  booktitle    = {International Conference on Learning Representations},
  year         = {2024}
}

\appendix
\renewcommand{\thetable}{A\arabic{table}}

\onecolumn 


\definecolor{codegray}{gray}{0.95}
\definecolor{commentgreen}{rgb}{0,0.6,0}
\definecolor{keywordblue}{rgb}{0.2,0.2,0.7}
\definecolor{stringred}{rgb}{0.58,0,0.1}
\definecolor{linenumbergray}{gray}{0.4}
\setcounter{secnumdepth}{2} 
\section{Details of Methods}

\subsection{LLM-based Abstract Operator: Prompt for Operator Generation}
\quad\textbf{Case-based Initial Operators Generation Prompt}
\vspace{1em}

\begin{lstlisting}[language=MyPython, style=vscodeLight, label={lst:case_based}]

GEN_CASE_OPS_PROMPT="""
You are a workflow design expert. The current {task_dtype} task is to extract workflow-related execution operations (execution operators) from specific problem descriptions and task resolutions. These execution operators are LLM-driven execution units.
- The operator's input includes contextual history and a functional description of the operator. The execution of this operator is ultimately implemented through the LLM (which may be supplemented by external tools).
- Extract key workflow nodes (e.g., execution, validation, etc.) that can be applied to similar tasks across different workflows.

The specific use case is as follows:
class Custom(object):
    def __init__(self, llm: AsyncLLM, name: str = "Custom"):
        self.llm = llm
        self.operator_prompt = "This is a common chat operator. "
    async def __call__(self, input):
        response = await self.llm(input, self.operator_prompt)
        return response

The final output format should be:
```python
class AsyncLLM:
    async def __call__(self, input, prompt):
        # Simulate an LLM response
        return "Processed: input with prompt: prompt"
        
class [operator_name](object):
    def __init__(self, llm: AsyncLLM=AsyncLLM(), name: str = "operator_name"):
        self.llm = llm
        self.operator_prompt = "[operator_prompt]" # only need string type's description for the operator function 
        # self.tool = tool_func() if the operator needs a tool.
    async def __call__(self, input):
        response = await self.llm(input, self.operator_prompt)
        # if the operator need tools, add the self.tool_func()
        # response = self.tool(response)
        return response

class workflow(object):
    def __init__(self,):
        self.llm = AsyncLLM()
        self.operator_list = [operator1, operator2, operator3, ...]
        
    async def __call__(self, problem):
        history = problem # cannot be adjusted
        for operator in self.operator_list[:-1]: 
            history += await operator(history) 
        reponse = await self.operator_list[-1](history)
        return reponse
```
\end{lstlisting}

\clearpage
\textbf{Operator Clustering and Preliminary Abstraction} 
\vspace{1em}

\begin{lstlisting}[language=MyPython, style=vscodeLight, label={lst:clustering}]
CLUSTER_OPS_PROMPT="""
Multiple task workflows contain execution operators. To derive clustered operators, you must:
- Cluster related/similar operators and prune non-essential ones
- Abstract generalized execution operations  (e.g., execution, validation, etc.)
- Ensure function names are concise, non-redundant, and minimize total operator count
- Express the solution as Python code
The specific use case is as follows:
class Custom(object):
    def __init__(self, llm: AsyncLLM, name: str = "Custom"):
        self.llm = llm
        self.operator_prompt = "This is a common chat operator. "
    async def __call__(self, input):
        response = await self.llm(input, self.operator_promp)
        return response
        
The final output format is:
```python
# the AsyncLLM code cannot be adjusted for simulation
class AsyncLLM:
    async def __call__(self, input, prompt):
        # Simulate an LLM response
        return "Processed: input with prompt: prompt"
        
class [operator_name](object):
    def __init__(self, llm: AsyncLLM, name: str = "operator_name"):
        self.llm = llm
        self.operator_prompt = "[operator_prompt]" # only need string type's description for the operator function 
        # self.tool = tool_func() if the operator needs a tool.
    async def __call__(self, input):
        response = await self.llm(input, self.operator_prompt)
        \""" if the operator need tools \"""
        # response = self.tool(response)
        return response
```
"""


\end{lstlisting}

\textbf{Deep Extraction for Abstract Execution Operators}
\vspace{1em}

\begin{lstlisting}[language=MyPython, style=vscodeLight, label={lst:abs_exc}]

DEEP_OPS_PROMPT="""
The input is some operations, but it is not abstract and general enough. 
- Deep merge and generate general execution operations (e.g., execution, validation, etc.), and is not more than two words as operation name. 
- Similar execution operations need to be merged to reduce the number of sub-operations. 
- It is required to be expressed in the form of Python code function. 

The specific use case is as follows:
class Custom(object):
    def __init__(self, llm: AsyncLLM, name: str = "Custom"):
        self.llm = llm
        self.operator_prompt = "This is a common chat operator. "

    async def __call__(self, input):
        response = await self.llm(input, self.operator_promp)
        return response
        
The final output format is:
```python
# the AsyncLLM code cannot be adjusted for simulation
class AsyncLLM:
    async def __call__(self, input, prompt):
        # Simulate an LLM response
        return "Processed: input with prompt: prompt"
        
class [operator_name](object):
    def __init__(self, llm: AsyncLLM, name: str = "operator_name"):
        self.llm = llm
        self.operator_prompt = "[operator_prompt]" # only need string type's description for the operator function 
        # self.tool = tool_func() if the operator needs a tool.
    async def __call__(self, input):
        response = await self.llm(input, self.operator_promp)
        \""" if the operator need tools \"""
        # response = self.tool(response)
        return response
```

DEEP_NEXT_PROMPT="Please determine if deep-merging can be continued and proceed to deep-merge and create general execution operations."
\end{lstlisting}

\vspace{2em}
\subsection{LLM-based Expansion: Prompt for LLM Optimizer}
\textbf{Workflow Optimize Prompt}
\vspace{1em}
\begin{lstlisting}[language=MyPython, style=vscodeLight, label={lst:prompt}]

WORKFLOW_OPTIMIZE_PROMPT = """
You are building a Graph and corresponding Prompt to jointly solve {type} problems. Referring to the given graph and prompt, which forms a basic example of a {type} solution approach, please reconstruct and optimize them. You can add, modify, or delete nodes, parameters, or prompts. Include your single modification in XML tags in your reply. Ensure they are complete and correct to avoid runtime failures. When optimizing, you can incorporate critical thinking methods like review, revise, ensemble (generating multiple answers through different/similar prompts, then voting/integrating/checking the majority to obtain a final answer), selfAsk, etc. Consider  Python's loops (for, while, list comprehensions), conditional statements (if-elif-else, ternary operators), or machine learning techniques (e.g., linear regression, decision trees, neural networks, clustering). The graph complexity should not exceed 10. Use logical and control flow (IF-ELSE, loops) for a more enhanced graphical representation.Ensure that all the prompts required by the current graph from prompt_custom are included.Exclude any other prompts. Output the modified graph and all the necessary Prompts in prompt_custom (if needed). The prompt you need to generate is only the one used in `prompt_custom.XXX` within Custom. Other methods already have built-in prompts and are prohibited from being generated. Only generate those needed for use in `prompt_custom`; please remove any unused prompts in prompt_custom. the generated prompt must not contain any placeholders. Considering information loss, complex graphs may yield better results, but insufficient information transmission can omit the solution. It's crucial to include necessary context during the process."""
\end{lstlisting}

\vspace{2em}
\textbf{Workflow Custom Use}
\vspace{1em}
\begin{lstlisting}[language=MyPython, style=vscodeLight, label={lst:workflow_cus}]

WORKFLOW_OPTIMIZE_PROMPT = """
Here's two examples of using the "op1, op2" operators in graph:
the first example is sequence processing:
```
history = problem # the workflow's memory
response = await self.op1(input=history) 
history += "op1: " + response['response']  
solution = await self.op2(input=history)  
```
the second example is parallel solution mechanism:
```
history = problem # the workflow's memory
for i in range(3):
    response = await self.op1(input=history) 
    history += f"op1, iter {i}: " + response['response']     
solution = await self.op2(input=history)  
```
\end{lstlisting}

\vspace{2em}
\subsection{Basic Structure of Workflow}
\quad\textbf{Workflow Structure}
\vspace{1em}
\begin{lstlisting}[language=MyPython, style=vscodeLight, label={lst:workflow}]

DatasetType = Literal["HumanEval", "MBPP", "GSM8K", "MATH", "HotpotQA", "DROP",
                      "alfworld_train", "alfworld_dev", "alfworld_test",
                      "textcraft_train", "textcraft_test"]

class Workflow:
    def __init__(
        self,
        name: str,
        llm_config,
        dataset: DatasetType,
    ) -> None:
        self.name = name
        self.dataset = dataset
        self.llm = create_llm_instance(llm_config)
        self.op = operator.op_xxx(self.llm)

    async def __call__(self, problem: str):
        """
        Implementation of the workflow with added processing and validation steps.
        """
        history = problem
        solution = await self.op(input=history)
        return solution['response'], self.llm.get_usage_summary()["total_cost"]
\end{lstlisting}
\vspace{1em}

\subsection{Abstract Operators}
\quad\textbf{ALFWorld operators}
\vspace{1em}

\begin{lstlisting}[language=MyPython, style=vscodeLight, label={lst:alfword}]

class Planner(object):
    def __init__(self, llm: AsyncLLM, name: str = "Planner"):
        self.llm = llm
        self.operator_prompt = (
            "Generate task plans and sequences including: "
            "1. Task decomposition\n"
            "2. Decision making\n"
            "3. Workflow management"
        )

    async def __call__(self, input):
        response = await self.llm(input, self.operator_prompt)
        return response


class Executor(object):
    def __init__(self, llm: AsyncLLM, name: str = "Executor"):
        self.llm = llm
        self.operator_prompt = (
            "Handle physical operations including: "
            "1. Spatial navigation\n"
            "2. Object manipulation\n"
            "3. Environment interaction"
        )

    async def __call__(self, input):
        response = await self.llm(input, self.operator_prompt)
        return response


class Validator(object):
    def __init__(self, llm: AsyncLLM, name: str = "Validator"):
        self.llm = llm
        self.operator_prompt = (
            "Perform verification tasks including: "
            "1. Progress assessment\n"
            "2. Quality control\n"
            "3. Error detection"
        )

    async def __call__(self, input):
        response = await self.llm(input, self.operator_prompt)
        return response
\end{lstlisting}
\vspace{2em}

\textbf{DROP Operators}
\vspace{1em}

\begin{lstlisting}[language=MyPython, style=vscodeLight, label={lst:drop_operators}]
class DataExtractor(object):
    def __init__(self, llm: AsyncLLM, name: str = "DataExtractor"):
        self.llm = llm
        self.operator_prompt = "Extract structured data including numerical values, categorical classifications, temporal information, and comparative relationships from input."
        
    async def __call__(self, input):
        response = await self.llm(input, self.operator_prompt)
        return response

class DataAnalyzer(object):
    def __init__(self, llm: AsyncLLM, name: str = "DataAnalyzer"):
        self.llm = llm
        self.operator_prompt = "Perform analytical operations including mathematical computations, temporal analysis, value comparisons, and logical validations on input data."
        
    async def __call__(self, input):
        response = await self.llm(input, self.operator_prompt)
        return response

class ResultProcessor(object):
    def __init__(self, llm: AsyncLLM, name: str = "ResultProcessor"):
        self.llm = llm
        self.operator_prompt = "Validate processed data and format results into final output structure with proper context and presentation."
        
    async def __call__(self, input):
        response = await self.llm(input, self.operator_prompt)
        return response
\end{lstlisting}
\vspace{1em}

\textbf{MATH Operators}
\vspace{1em}

\begin{lstlisting}[language=MyPython, style=vscodeLight, label={lst:MATH_Operators}]
class Analyzer(object):
    def __init__(self, llm: AsyncLLM, name: str = "Analyzer"):
        self.llm = llm
        self.operator_prompt = "Analyze input to identify components, constraints, relationships and requirements."
    
    async def __call__(self, input):
        response = await self.llm(input, self.operator_prompt)
        return response

class Processor(object):
    def __init__(self, llm: AsyncLLM, name: str = "Processor"):
        self.llm = llm
        self.operator_prompt = "Process, transform and compute solutions through extraction, structuring and logical reasoning."
    
    async def __call__(self, input):
        response = await self.llm(input, self.operator_prompt)
        return response

class Validator(object):
    def __init__(self, llm: AsyncLLM, name: str = "Validator"):
        self.llm = llm
        self.operator_prompt = "Validate solution correctness, completeness and format the final output appropriately."
    
    async def __call__(self, input):
        response = await self.llm(input, self.operator_prompt)
        return response
\end{lstlisting}
\vspace{1em}

A list of all abstract operator names is provided below:
\begin{table*}[htpb]
  \centering
  \setlength{\tabcolsep}{4mm}{
  \begin{tabular}{c c }
    \toprule
     \textbf{Benchmark} & Operators  \\
     \midrule
      HotpotQA  & InfoExtractor, ResponseGenerator, ContentValidator    \\
      DROP  & DataExtractor, DataAnalyzer, ResultProcessor  \\
      HumanEval  & AnalyzePlan, BuildExecute, ValidateCheck   \\
      MBPP  & CodeProcessor, TestManager, SolutionEngine   \\
      GSM8K  & InfoExtractor, MathProcessor, SolutionChecker, OutputFormatter  \\
      MATH  & Analyzer, Processor, Validator   \\
      ALFWorld  & Planner, Executor, Validator   \\
      TextCraft  & WorkflowPlanner, ResourceManager, TaskExecutor, QualityVerifier  \\
      \bottomrule
  \end{tabular}
  }
  \label{tab:ops_list}
\end{table*}

\subsection{Mapping Workflow from Operators to Code}
\vspace{1em}
\quad\textbf{Example of Workflow}
\vspace{1em}
\begin{lstlisting}[language=MyPython, style=vscodeLight, label={lst:exp_workflow}]
    async def __call__(self, problem: str):
        """
        Implementation of the workflow with enhanced problem breakdown and verification
        """
        history = problem
        
        # Stage 1: Problem breakdown and key information extraction
        breakdown_prompt = f"Break down this math problem step by step. Identify: 1) What is being asked, 2) All given quantities and their units, 3) What calculations are needed, 4) Any potential misinterpretations to avoid.\n\nProblem: {problem}"
        breakdown = await self.InfoExtractor(input=breakdown_prompt)
        history += "\n\nPROBLEM BREAKDOWN:\n" + breakdown['response']
        
        # Stage 2: Initial solution with detailed steps
        solution = await self.MathProcessor(input=history)
        history += "\n\nINITIAL SOLUTION:\n" + solution['response']
        
        # Stage 3: Enhanced solution verification with specific error checking
        verification_prompt = f"{history}\n\nCarefully verify this solution by:\n1) Recalculating each arithmetic step independently\n2) Checking if all given numbers from the problem are used correctly\n3) Verifying unit conversions and final units\n4) Confirming the final answer makes logical sense\n\nIf you find ANY calculation errors or missing steps, clearly state 'ERRORS FOUND' and explain what needs correction."
        verification = await self.SolutionChecker(input=verification_prompt)
        history += "\n\nVERIFICATION:\n" + verification['response']
        
        # Stage 4: Conditional correction if errors were found
        if "ERRORS FOUND" in verification['response'].upper() or "ERROR" in verification['response'].upper():
            correction_prompt = f"{history}\n\nSince errors were identified, recalculate the solution from scratch. Show each calculation step clearly and double-check your arithmetic."
            corrected_solution = await self.MathProcessor(input=correction_prompt)
            history += "\n\nCORRECTED SOLUTION:\n" + corrected_solution['response']
        
        # Stage 5: Final answer with clear formatting
        final_prompt = history + "\n\nBased on all the work above, provide the final numerical answer. Ensure it's a single number and use the format: FINAL ANSWER: [number]"
        final_solution = await self.MathProcessor(input=final_prompt)
        
        return final_solution['response'], self.llm.get_usage_summary()["total_cost"]
\end{lstlisting}
\vspace{1em}
In this example,
\begin{itemize}
\item self.MathProcessor, self.InfoExtractor and self. SolutionChecker are key operators in the workflow. At different stages, operators also appear and execute again to improve performance. 
\item Every operator has only one input and one output as basic program units. LLM optimizer generates code to build the flow of all input/output variables between Nodes and Operators. To get better workflow context information, embed the history block into the workflow to store previous operators' responses.
\item The operators are the interface for building nodes. The optimizer uses them to generate/modify prompts, which are appended to the history and used by the next operator.
\end{itemize}
\vspace{1em}


\section{Dataset and Cost-Performance}\label{sec:DATASET_AND_PARETO_FRONT}
\subsection{The Dataset Statistics}
We divide the data into validation and test sets using a 1:4 ratio for each benchmark dataset. Specifically, for the embodied and game tasks, we select the part training dataset that satisfies the validation ratio. 
Last, the ALFWorld task is 33 samples, and the TextCraft task is 20 samples in the validation dataset. Since embodied tasks and game tasks involve interaction with the environment, we set the workflow optimization rounds to 10 and other tasks to 20.

\subsection{Cost Analysis}
\noindent Detailed cost-performance benchmarks for DROP under varied executors reveal significant efficiency improvements. We select the top three workflows from the validation set for detailed analysis. A\textsuperscript{2}Flow integrated with GPT-4o outperformed AFLOW by 1.62\% in accuracy while consuming only 51.37\% of its computational resources (48.63\% cost reduction). For DeepSeek-v3, A\textsuperscript{2}Flow achieved equivalent efficacy at 43.02\% of AFLOW's operational expense in the optimal workflow configuration.

\begin{table*}[htpb]
  \centering
  \setlength{\tabcolsep}{4mm}{
  \begin{tabular}{c c c c c}
    \toprule
     \textbf{Model} & Method & Workflow & Socre(\%) & Cost(\$) \\
     \midrule
      gpt-4o-mini  & AFLOW & Top-1 & 0.8200 & 1.3739   \\
      gpt-4o-mini  & AFLOW & Top-2 & 0.8145 & 0.4162  \\
      gpt-4o-mini  & AFLOW & Top-3 & 0.8051 & 0.8196   \\
      gpt-4o-mini  & Our & Top-1 & 0.8575 & 0.5130   \\
      gpt-4o-mini  & Our & Top-2 & 0.8321 & 0.2638  \\
      gpt-4o-mini  & Our & Top-3 & 0.8311 & 0.3877   \\

      \midrule
      gpt-4o  & AFLOW & Top-1 & 0.8777 & 17.2492   \\
      gpt-4o  & AFLOW & Top-2 & 0.8761 & 5.3946  \\
      gpt-4o  & AFLOW & Top-3 & 0.8606 & 11.2064   \\
      gpt-4o  & Our & Top-1 & 0.8920 & 8.3886   \\
      gpt-4o  & Our & Top-2 & 0.8245 & 4.5417  \\
      gpt-4o  & Our & Top-3 & 0.8279 & 6.5881   \\

    \midrule
      deepseek-v3  & AFLOW & Top-1 & 0.8903 & 2.1612   \\
      deepseek-v3  & AFLOW & Top-2 & 0.8816 & 0.6922  \\
      deepseek-v3  & AFLOW & Top-3 & 0.8465 & 1.3568   \\
      deepseek-v3  & Our & Top-1 & 0.8921 & 0.9298   \\
      deepseek-v3  & Our & Top-2 & 0.8527 & 0.4567  \\
      deepseek-v3  & Our & Top-3 & 0.8356 & 0.7079   \\
      \bottomrule
  \end{tabular}
  \vspace{1em}
  }
  \label{tab:appendix_cost_analyse}
\end{table*}


\section{Case Study}\label{sec:CASE_STUDY}
\subsection{The optimization process for the embodied task}
Using the ALFWorld benchmark as an example, we demonstrate how A\textsuperscript{2}Flow iteratively improves workflows based on the abstract operators through the search process.

\vspace{1em}

\begin{lstlisting}[language=MyPython, style=vscodeLight, label={lst:opt_process}]
{
    "1": {
        "score": 0.21212121212121213,
        "success": {
            "3": {
                "modification": "Added Validator operator after Executor to verify the solution and provide feedback, and enhanced the Executor prompt to be more specific for ALFWorld environment tasks",
                "score": 0.24242424242424243
            },
            "2": {
                "modification": "Add Planner operator before Executor to create a two-stage workflow: planning then execution",
                "score": 0.30303030303030304
            }
        },
        "failure": {
            "6": {
                "modification": "Enhanced the Executor with an iterative self-refinement approach, where the Executor processes the problem multiple times with accumulated context, and improved the prompt to be more specific for ALFWorld environment tasks with detailed action formatting requirements.",
                "score": 0.09090909090909091
            }
        }
    },
    "2": {
        "score": 0.30303030303030304,
        "success": {
            "7": {
                "modification": "Add iterative planning refinement by having the Executor provide feedback to the Planner for plan improvement, with a maximum of 2 iterations to prevent infinite loops",
                "score": 0.3333333333333333
            }
        },
        "failure": {
            "5": {
                "modification": "Add Validator operator after Executor to verify execution results and provide quality control feedback",
                "score": 0.15151515151515152
            }
        }
    },
    "3": {
        "score": 0.24242424242424243,
        "success": {},
        "failure": {
            "4": {
                "modification": "Add Planner operator before Executor to create a structured planning phase that decomposes the task before execution",
                "score": 0.15151515151515152
            }
        }
    },
    "7": {
        "score": 0.3333333333333333,
        "success": {},
        "failure": {
            "10": {
                "modification": "Add a strategic validation checkpoint after the iterative refinement loop but before final execution, using the Validator operator to analyze the accumulated history and provide strategic insights for the final execution phase.",
                "score": 0.15151515151515152
            }
            "9": {
                "modification": "Replace the iterative refinement loop with a parallel execution strategy that generates multiple execution approaches simultaneously, then consolidates them. This avoids the failed three-stage validation pattern while utilizing the Executor more effectively through diverse strategy generation.",
                "score": 0.21212121212121213
            },
            "8": {
                "modification": "Add Validator operator integration after each execution attempt to verify progress and detect potential issues before plan refinement, creating a three-stage iterative process (Plan -> Execute -> Validate)",
                "score": 0.12121212121212122
            }
        }
    }
}
\end{lstlisting}
\vspace{1em}

\noindent \textbf{Successful Optimization Steps:}
\begin{itemize}
\item  Round 1 $\rightarrow$ Round 2 (Score improved from 0.2121 to 0.3030): The key optimization was the addition of the Planner operator, relying solely on a two-
stage workflow: planning + execution, which improved the workflow's performance.
\item  Round 2 $\rightarrow$  Round 7 (Score improved from 0.3030 to 0.3333): The key optimization was the integration of iterative planning refinement—this critical enhancement not only bolstered solution robustness but also facilitated embodied task planning and executable action.
\end{itemize}

\noindent \textbf{Less Effective Optimization Steps:}
\begin{itemize}
\item  Round 1 $\rightarrow$ Round 6 (Score decreased from 0.2121 to 0.0909): The key change was the addition of the Validator operator. The reliance on self-model judgments, as opposed to real-world environmental feedback, introduces evaluator bias that degrades task executing performance.
\item  Round 7 $\rightarrow$ Round 9 (Score decreased from 0.3333 to 0.1515): The key change was replacing the iterative refinement loop with a parallel execution strategy that generates multiple execution approaches simultaneously, then consolidating them. Facing an open-ended problem of significant complexity,  its initial long chain-of-thought approach, reliant on robustness-centric design, ultimately generated ungrounded decisions due to overthinking and error accumulation.
\end{itemize}

\section{More Details of Ablation Study}
In the MATH benchmark, all abstract operators’ names are as follows:
\begin{table*}[htpb]
  \centering
  \setlength{\tabcolsep}{4mm}{
  \begin{tabular}{p{6cm} p{9cm}}

    \toprule
     \textbf{Operator Type} & Operators  \\
     \midrule
      Initial Operators & InfoExtractor, ResponseGenerator, ContentValidator, MathBoundaryIdentifier, EstimationValidator, ParticipantIdentificationOperator, FractionMathOperator, FractionConversionOperator, FactorFinder, CoprimeGenerator, TotalCounter, MathFormulaProcessor, MathCalculationOperator, MoneyRoundingOperator, ParseStemLeafPlot, CalculateStats, ConstraintSolver...    \\
      \hline
      Operator Clustering  & ProblemAnalyzer, DataProcessor, MathematicalSolver, LogicalReasoner, SolutionValidator, OutputFormatter  \\
      \hline
      Deep Extraction Operators-step1  & AnalyzeProblem, ProcessData, ComputeLogic, ValidateSolution, FormatOutput  \\
      Deep Extraction Operators-step2  & AnalyzeInput, ProcessTask, FinalizeOutput  \\
      Deep Extraction Operators-step3  & Analyze, Process, Finalize  \\
      \hline
      Deep Extraction Operators-merge  & Analyzer, Processor, Validator  \\

      \bottomrule
  \end{tabular}
  }
    \caption{Ablation study for three components: initialization operators, clustering operators and deep execution operators. We select a part of the initial operators and the first reasoning path to present the deep merging process.}
  \label{tab:all_operators_for_math}
\end{table*}

\vspace{1em}

\twocolumn
\end{document}